\documentclass{article}
\usepackage[T1]{fontenc}
\usepackage[utf8]{inputenc}
\usepackage{lmodern}
\usepackage[english]{babel}
\usepackage{amsmath}
\usepackage{amssymb}
\usepackage{amsthm}
\usepackage{epigraph}
\usepackage{graphicx}
\usepackage{cite}
\usepackage{hyperref}

\begin{document}
\title{The Computational Power of Dynamic Bayesian Networks}
\author{Joshua Brulé\footnote{Department of Computer Science, University of Maryland, College Park, MD 20742. \href{mailto:jbrule@cs.umd.edu}{jbrule@cs.umd.edu}}}
\date{}
\maketitle

\begin{abstract}
This paper considers the computational power of constant size, dynamic Bayesian networks. Although discrete dynamic Bayesian networks are no more powerful than hidden Markov models, dynamic Bayesian networks with continuous random variables and discrete children of continuous parents are capable of performing Turing-complete computation. With modified versions of existing algorithms for belief propagation, such a simulation can be carried out in real time. This result suggests that dynamic Bayesian networks may be more powerful than previously considered. Relationships to causal models and recurrent neural networks are also discussed.
\end{abstract}

\section{Introduction}

Bayesian networks are probabilistic graphical models that represent a set of random variables and their conditional dependencies via a directed acyclic graph. Explicitly modeling the conditional dependencies between random variables permit efficient algorithms to perform inference and learning in the network. Causal Bayesian networks have the additional requirement that all edges in the network model a causal relationship.

Dynamic Bayesian networks are the time-generalization of Bayesian networks and relate variables to each other over adjacent time steps. Dynamic Bayesian networks unify and extend a number of state-space models including hidden Markov models, hierarchical hidden Markov models and Kalman filters. Dynamic Bayesian networks can also be seen as the natural extension of acyclic causal models to models that permit cyclic causal relationships, while avoiding problems with causal models that try to model temporal relationships with an atemporal description \cite{Poole&Crowley2013}.

A natural question is what is the expressive power of such networks. The result in this paper shows that although discrete dynamic Bayesian networks are sub-Turing in computational power, introducing continuous random variables with discrete children is sufficient to model Turing-complete computation. In addition, the distributions used in the construction are such that the marginal posterior probabilities of random variables in the network can be effectively computed with modified versions of existing algorithms. Ignoring the overhead from arbitrary precision arithmetic, the simulation can be conducted with only a constant time penalty.

\section{The Model and Main Results}

A Bayesian network \cite{Pearl1985} consists of a directed-acyclic graph, $G$ over a set $\mathbf{V} = \{ V_1, \ldots, V_n \}$ of vertices and a probability distribution $P(\mathbf{v})$ over the set of variables that correspond to the vertices in $G$. A Bayesian network ``factorizes'' the probability distribution over its variables, by requiring that each variable, $v_i$, is conditionally independent of its non-descendants, given its parents (denoted $pa(v_i)$). This is the Markov condition \cite{Bareinboim2012}:

\[
    P(x_1, \ldots ,x_n) = \prod_{i}P(x_i | pa_i)
\]

Dynamic Bayesian networks (DBN) extend Bayesian networks to model a probability distribution over a semi-infinite collection of random variables, with each collection of random variables modeling the system at a point in time \cite{Dean&Kanazawa1989}. Following the conventions in \cite{Murphy2002}, the collections are denoted $Z_1, Z_2, \ldots$ and variables are partitioned $Z_t = (U_t, X_t, Y_t)$ to represent input, hidden and output variables of a state space model. Such a network is ``dynamic'' in the sense that it can model a dynamic system, not that the network topology changes over time.

A DBN is defined as a pair $(B_1, B_\rightarrow)$, where $B_1$ is a Bayesian network that defines the prior $P(Z_1)$ and $B_\rightarrow$ is a two-slice temporal Bayes net (2TBN) that defines $P(Z_t | Z_{t-1})$ via a directed acyclic graph:

\[
    P(Z_t | Z_{t-1}) = \prod_{i=1}^N P(Z_t^i | pa(Z_t^i))
\]
where $Z_t^i$ is the $i^{\text{th}}$ node at time $t$, and $pa(Z_t^i)$ are the parents of $Z_t^i$ in the graph. The parents of a node can either be in the same time slice or in the previous time slice (i.e. the model is first-order Markov).

The semantics of a DBN can be defined by ``unrolling'' the 2TBN until there are $T$ time-slices; the joint distribution is then given by:

\[
    P(Z_{1:T}) = \prod_{t=1}^T \prod_{i=1}^N P(Z_T^i | pa(Z_t^i))
\]

Analyzing the computational power of a DBN requires defining what it means for a DBN to accept (and halt) or reject an input. Define an input sequence, $\{U_{t}\}$ of Bernoulli random variables to model the (binary) input. Similarly, define an output sequence $\{Y_{t}\}$ ($Y_t \in \{run, halt_{0}, halt_{1} \}$) to represent whether the machine has halted and the answer that it gives. Given an input, $in_1, in_2, \ldots, in_t$, to a decision problem, the machine modeled by the DBN to has halted and accepted at time $t$, if and only if $P(Y_t=halt_{1} | U_{1} = in_1, \ldots, U_{n} = in_t) > 0.5$ and halted and rejected if and only if $P(Y_t=halt_{0} | U_{1} = in_1, \ldots, U_{n} = in_t) > 0.5$.

\subsection{Discrete dynamic Bayesian networks are not Turing-complete}

``Discrete'' Bayesian networks are Bayesian networks where all random variables have some finite number of outcomes, i.e. Bernoulli or categorical random variables. If dynamic Bayesian networks are permitted to increase the number of random variables over time, then simulating a Turing-machine becomes trivial: simply add a new variable each time step to model a newly reachable cell on the Turing machine's tape. However, this requires some `first-order' features in the language used to specify the network and the computational effort required at \emph{each step} of the simulation will grow without bound.

With a fixed number of random variables at each time step and the property that DBNs are first-order Markov, the computational effort per step remains constant. However, discrete DBNs have sub-Turing computational power. Intuitively, a discrete DBN cannot possibly simulate a Turing machine since there is no way to store the contents of the machine's tape.

More formally, any discrete Bayesian network can be converted into a hidden Markov model \cite{Murphy2002}. This is done by `collapsing' the hidden variables ($X_t$) of the DBN into a single random variable by taking the Cartesian product of their sample space. The `collapsed' DBN models a probability distribution over a exponentially larger, but still finite sample space. Hidden Markov models are equivalent to probabilistic finite automata \cite{Dupont2005} which recognize the stochastic languages. Stochastic languages are in the RP-complexity class and thus discrete DBNs are not Turing complete.

\subsection{A dynamic Bayesian network with continuous and discrete variables}

A 2TBN can be constructed to simulate the transitions of a two stack push-down automaton (PDA), which is equivalent to the standard one tape Turing machine. A two stack PDA consists of a finite control, two unbounded binary stacks and an input tape. At each step of computation, the machine reads and advances the input tape, reads the top element of each stack and can either push a new element, pop the top element or leave each stack unchanged. The state of the control can change as function of previous state and the read symbols. When the control reaches one of two possible halt states ($\{halt_0, halt_1\}$), the machine stops and its output to the decision problem it was computing is defined which of the halt states it stops on.

A key part of the construction is using a Dirac distribution to simulate a stack. A Dirac distribution centered at $\mu$ can be defined as the limit of normal distributions:

\[
    \delta(\mu) \equiv \lim_{\sigma \downarrow 0} \frac{1}{\sigma \sqrt{2 \pi}} e^{\frac{-x^2}{2 \sigma^2}}
\]

A single Dirac distributed random variable is sufficient to simulate a stack. The stack construction adapted from \cite{siegelmann1995} encodes a binary string $\omega = \omega_1\omega_2\ldots\omega_n$ into the number:

\[
    q = \sum_{i=1}^n \frac{2 \omega_i + 1}{4^i}
\]

Note that if the string begins with the value 1, then $q$ has a value of at least $3/4$ and if the string begins with $0$, then $q$ is less than $1/2$ - there is never a need to distinguish among two very close numbers to read the most significant digit. In addition, the empty string is encoded as $q = 0$, but any non-empty string has value at least $1/4$.

All random variables, except for the stack random variables, are categorically distributed - thus, the conditional probabilities densities between them can be represented using standard conditional probability tables.

Extracting the top value from a stack requires a conditional probability distribution for a Bernoulli random variable ($Top \in \{0, 1\}$), given a Dirac ($Stack \in \mathbb{R}$) distributed parent. The Heavyside step function meets this requirement and is defined as the limit of logistic functions (or, more generally, softmax functions), centered at $1/2$:

\[
    H(x) \equiv \lim_{k \rightarrow \infty} \frac{1}{1 + e^{-k(x - 1/2)}}
\]

The linear operation $4q - 2$ transfers the range of $q$ to at least $1$ when the top element of the stack is $1$ and no more than $0$ when the top element of the stack is 0. Then, the conditional probability density function:

\[
    P(Top | Stack=q) = H(4q - 2)
\]
yields $P(Top) = 1$ whenever the top element of the stack is $1$ and $P(Top) = 0$ whenever the top element of the stack is $0$.

Similarly, a conditional probability distribution can be defined for Bernoulli random variable $Empty \in \{0, 1\}$, as:

\[
P(Empty | Stack=q) = 1 - H(4q)
\]
to check if a stack is empty.

Finally, the linear operations $\frac{q}{4} + \frac{2b+1}{4}$ and $4q - (2b+1)$ push and pop $b$, respectively, from a stack. The conditional probability density for a stack at time $t+1$, given a stack at time $t$, the top of the stack at time $t$, and action to be performed on the stack ($Action_t \in \{push_0, push_1, pop, noop\}$) is fully described as follows:

\[
    P(Stack_{t+1} | Top_t=p, Stack_t=q, Action_t=push_0) = \delta(q/4 + 1/4)
\]
\[
    P(Stack_{t+1} | Top_t=p, Stack_t=q, Action_t=push_1) = \delta(q/4 + 3/4)
\]
\[
    P(Stack_{t+1} | Top_t=p, Stack_t=q, Action_t=pop) = \delta(4q - (2p + 1))
\]
\[
    P(Stack_{t+1} | Top_t=p, Stack_t=q, Action_t=noop) = \delta(q)
\]

Since there are two stacks in the full construction, they are labeled, at time $t$, as $Stack_{a, t}$ and $Stack_{b, t}$. The rest of the construction is straightforward. $State_{t}$, $Action_a$ and $Action_b$ are functions of $State_{t-1}, Top_{a,t}, Empty_{a, t}, Top_{b, t}, Empty_{b, t}$ and $in_t$. Since all of these are discrete random variables, the conditional probability densities is simply the transition function of the PDA, written as a (0, 1) stochastic matrix. As expected $P(Y=halt_i | State) = 1$ if $State$ is that halt state, and $0$ otherwise.

Finally, the priors for the dynamic Bayesian network are simply $P(Stack_{a,1}) = P(Stack_{b,1}) = \delta(0)$, $P(State_1 = q_0) = 1$, where $q_0$ is the initial state.

As described, this construction is somewhat of an abuse of the term `probabilistic graphical model' - all probability mass is concentrated into a single event for every random variable in the system, for every time step. However, it is easy to see this construction faithfully simulates a two stack machine, as each random variable in the construction corresponds exactly to a component of the simulated automaton.

\subsection{Exact inference in continuous-discrete Bayesian networks}

This construction requires continuous random variables, which raise concerns as to whether the marginal posterior probabilities can be effectively computed. The original junction tree algorithm \cite{Lauritzen1988} and cut-set conditioning \cite{Pearl1988} approaches to belief propagation compute exact marginals for arbitrary DAGs, but require discrete random variables. Lauritzen's algorithm \cite{Lauritzen1992} conducts inference in mixed graphical models, but is limited to conditional linear Gaussian (CLG) continuous random variables. In a CLG model, let $X$ be a continuous node, $\mathbf{A}$ be its discrete parents, and $Y_1, \ldots, Y_k$ be continuous parents. Then

\[
    p(X | \mathbf{a}, \mathbf{y}) = N(\mathbf{w}_{\mathbf{a},0} + \sum_{i=1}^k w_{\mathbf{a}, i} y_i ; \sigma_{\mathbf{a}}^2 )
\]

Lauritzen's algorithm can only conduct approximate inference, since the true posterior marginals may be some multimodal mix of Gaussians, while the algorithm itself only supports CLG random variables. However, the algorithm is exact in the sense that it computes exact first and second moments for the posterior marginals which is sufficient for the Turing machine simulation.

Laurientz's algorithm does not permit discrete random variables to be children of continuous random variables. Lerner's algorithm \cite{Lerner2001} extends Lauritzen's algorithm to support softmax conditional probability densities for discrete children of continuous parents. Let A be a discrete node with the possible values $a_1, \ldots, a_m$ and let $Y_1, \ldots, Y_k$ be its parents. Then:

\[
    P(A = a_i | y_1, \ldots, y_k) = \frac{exp(b^i + \sum_{l=1}^n w_l^i y_l)}{\sum_{j=1}^m exp(b^j + \sum_{l=1}^n w_l^i y_l) }
\]

Like Lauritzen's algorithm, Lerner's algorithm computes approximate posterior marginals - relying on the observation that the product of a softmax and a Gaussian is approximately Gaussian - but exact first and second moments, up to errors in the numerical integration used to compute the best Gaussian approximation of the product of a Gaussian and a softmax. This calculation is actually simpler in the case where the softmax is replaced with a Heavyside and the Lerner algorithm can run essentially unmodified with a mixture of Heavyside and softmax conditional probability densities. In the case of Dirac-distributed parents, with Heavyside conditional probability densities, numeric integration is unnecessary and no errors are introduced in computing the first and second moments of the posterior distribution.

Any non-zero variance for the continuous variables will `leak' probability to other values for the `stack' random variables in the Turing machine simulation, eventually leading to errors. Lauritzen's original algorithm assumes positive-definite covariance matrices for the continuous random variables, but can be extend to handle degenerate Gaussians \cite{raphael2003}. In summary: posterior marginals for the Turing machine simulation can be computed exactly, using a modified version of the Lerner algorithm when restricted to Dirac distributed continuous random variables with Heavside conditional probability densities. If Gaussian random variables and softmax conditional probability densities are also introduced, then the first and second moments of the posterior marginals can be computed `exactly', up to errors in numerical integration, although this will slowly degrade the quality of the Turing machine simulation in later time steps.

Inference in Bayesian networks is NP-hard \cite{cooper1990}. However, assuming that arithmetic operations can be computed in unit time over arbitrary-precision numbers (e.g. the real RAM model), the work necessary at each time step is constant. Thus, dynamic Bayesian networks can simulate Turing-machines with only a constant time overhead in the real RAM model, and slowdown proportional to the time complexity of arbitrary precision arithmetic otherwise.

\section{Discussion}

This result suggests that causal Bayesian networks may be a richer language for modeling causality than currently appreciated. Halpern \cite{Halpern2000} suggests that for general causal reasoning, a richer language, including some-first order features may be needed. First-order features will likely be very useful for causal modeling in practice, but the Turing-complete power of dynamic Bayesian networks suggests that first-order features may be unnecessary.

This result for dynamic Bayesian networks is analogous to Siegelmann and Sontag's proof that a recurrent neural network can simulate a Turing machine in real time \cite{siegelmann1995}. In fact, neural networks and Bayesian networks turn out to have very similar expressive power:

\begin{enumerate}
    \item Single perceptron $\approx$ Gaussian naive Bayes (Logistic regression) \cite{ng2001}
    \item Multilayer perceptron $\approx$ Full Bayesian network (Universal function approximation) \cite{cybenko1989} \cite{varando2014}
    \item Recurrent neural network $\approx$ Dynamic Bayesian network (Turing complete)
\end{enumerate}

There is an interesting gap in decidability - it takes very little to turn a sub-Turing framework for modeling into a Turing-complete one. In the case of neural networks, a single recurrent layer, with arbitrary-precision rational weights and a saturating linear transfer function is sufficient. With dynamic Bayesian networks, two time-slices, continuous-valued random variables with a combination of linear and step function conditional probability densities is sufficient.

Although such a simple recurrent neural network is theoretically capable of performing arbitrary computations, practical extensions include higher-order connections, \cite{pineda1988}, `gates' in long short-term memory \cite{hochreiter1997}, and even connections to an `external' Turing machine \cite{graves2014}. These additions enrich the capabilities of standard neural networks and make it easier to train them for complex algorithmic tasks.

An interesting question is to what degree dynamic Bayesian networks can be similarly extended and how the `core' dynamic Bayesian network being capable of Turing-complete computation affects the overall performance of such networks.

\section*{Acknowledgements}

I would like to thank James Reggia, William Gasarch and Brendan Good for their discussions and helpful comments on early drafts of this paper.

\bibliography{dbn}{}
\bibliographystyle{ieeetr}

\end{document}